\documentclass[sigconf,nonacm]{acmart}
\usepackage{tikz}
\usepackage{standalone}
\usepackage{graphicx}
\usepackage{enumitem}
\usepackage{subcaption}  
\usepackage[normalem]{ulem}  
\usepackage{xcolor} 
\AtBeginDocument{%
  }

\setcopyright{acmlicensed}
\copyrightyear{2025}
\acmYear{2025}
\acmDOI{XXXXXXX.XXXXXXX}
\acmISBN{978-1-4503-XXXX-X/2018/06}

\usepackage{multirow}
\usepackage{booktabs}

\begin{document}

\title{Beyond Attention: Learning Spatio-Temporal Dynamics with Emergent Interpretable Topologies}

\author{Sai Vamsi Alisetti}
\authornote{Both authors contributed equally to this research.}
\email{saivamsi@ucsb.edu}
\affiliation{%
  \institution{University of California, Santa Barbara}
  \city{Santa Barbara}
  \state{California}
  \country{USA}
}

\author{Vikas Kalagi}
\authornotemark[1]
\email{vikaskalagi@ucsb.edu}
\affiliation{%
  \institution{University of California, Santa Barbara}
  \city{Santa Barbara}
  \state{California}
  \country{USA}
}

\author{Sanjukta Krishnagopal}
\email{sanjukta@ucsb.edu}
\affiliation{%
  \institution{University of California, Santa Barbara}
  \city{Santa Barbara}
  \state{California}
  \country{USA}
}

\begin{abstract}
Spatio-temporal forecasting is critical in applications such as traffic prediction, energy demand modeling, and weather monitoring. While Graph Attention Networks (GATs) are popular for modeling spatial dependencies, they rely on predefined connectivity and dynamic attention scores, introducing inductive biases and computational overhead, often at the cost of interpretability.

We propose \textbf{InterGAT}, a simplified alternative to GAT that replaces masked attention with a fully learnable, symmetric node interaction matrix, capturing latent spatial relationships without relying on fixed graph topologies. 
Our framework, \textbf{InterGAT-GRU}, which incorporates a GRU-based temporal decoder, outperforms the baseline GAT-GRU in forecasting accuracy, achieving at least a 21\% improvement on the SZ-Taxi dataset and a 6\% improvement on the Los-Loop dataset across all forecasting horizons (15 to 60 minutes). Additionally, we observe a reduction in training time by 60-70\% compared to GAT-GRU baseline.

Interestingly, the learned interaction matrix reveals interpretable structure: it recovers sparse, topology-aware attention patterns that align with community structure. Spectral and clustering analyses indiate that the model captures both localized and global dynamics, offering insights into the emergent topology driving its predictions. This highlights how structure learning can simultaneously enhance predictive performance, efficiency, and topological interpretability in dynamic spatio-temporal graph domains.
\end{abstract}






\renewcommand\footnotetextcopyrightpermission[1]{}  
\settopmatter{printacmref=false, printccs=false, printfolios=true}

\maketitle


\section{Introduction}

Spatio-temporal forecasting is central to many real-world systems, from intelligent transportation networks~\cite{zhao2019t}, to energy grids~\cite{wang2024robust}, and environmental monitoring~\cite{xie2023spatio}. It also plays a vital role in healthcare~\cite{wang2022deepcovidnet}, climate science~\cite{ravuri2021skilful}, agriculture~\cite{shahhosseini2021forecasting}, urban planning~\cite{li2023urbanflow}, and prediction of social events~\cite{wu2023learning}, making it a foundational task in domains where spatial and temporal dependencies are key. These tasks involve modeling both the spatial dependencies between entities (e.g., road sensors) and the temporal evolution of their signals over time. Graph Neural Networks (GNNs) \cite{4700287} have become the de facto tools for spatial modeling in such domains. 

Graph Attention Networks (GATs) \cite{velivckovic2018gat}, in particular, stand out due to their ability to adaptively weight neighboring nodes through attention mechanisms.
However, conventional GATs have two limitations: 1) they rely on predefined adjacency matrices that encode a fixed graph structure and 2) compute pairwise attention scores at every forward pass. These design choices introduce structural biases, add computational complexity, and often limit interpretability. 
Previous works have attempted to address this limitation by introducing adaptive or dynamic adjacency matrices \cite{zhang2021stgsl}, \cite{bai2020adaptive}, \cite{fang2021spatial}, \cite{xu2023timegnn}, which allow connectivity patterns to evolve based on node features, tasks, or temporal contexts. 
Yet these approaches typically retain costly attention computations and require recomputing the graph structure at every time step.

In this work, we propose an efficient and interpretable simplification of this paradigm. Instead of relying on masked attention over a fixed graph, we introduce a \textit{ fully learnable symmetric node-to-node interaction matrix} that captures latent structural relationships directly from data. This matrix is shared across time and trained end-to-end, allowing the model to discover meaningful topological patterns such as modular communities or long-range dependencies without access to prior structural knowledge.

This design enables not only improved training efficiency but also topology-aware interpretability. By analyzing the learned interaction matrix, we can gain insights into which node relationships are most influential for forecasting, and how attention is distributed across the graph. Spectral analyses reveal that the learned interaction matrix exhibits a full smooth eigenvalue spectrum, and the most predictive modes capture sharp contrasts across key node pairs. These findings echo trends in recent works that investigate the spectral properties of attention mechanisms in graphs \cite{brody2021attentive, kim2022pure} and their alignment with underlying topological structure \cite{ying2019gnnexplainer, dwivedi2023graph}

Our community alignment analyses reveal that different heads of the model specialize to focus attention on localized and global structure respectively. The learned attention often align with topology-aware community structure, bridging interpretable deep learning and classical network science. To model temporal evolution, we pair this spatial module with a GRU-based temporal decoder, yielding InterGAT-GRU—a spatio-temporal forecasting framework.

This architecture offers several advantages:
\begin{sloppypar}
\begin{itemize}[leftmargin=*, noitemsep, topsep=0pt, partopsep=0pt, parsep=0pt]
    \item \textbf{Efficiency:} It improves computational efficiency across time horizons. Specifically, it reduced training time by 60-70\% compared to an equivalent GAT-GRU baseline for the SZ-Taxi and Los-Loop dataset in the 15 minute horizon. These comparisons are made using identical hidden layer dimensions, GRU decoders, and identical hardware settings.
    \item \textbf{Performance:} The model consistently outperforms equivalent GAT-GRU baselines in prediction on two benchmark traffic datasets across all forecasting horizons.
    \item \textbf{Interpretability:} It learns symmetric, sparse interaction matrices that uncover modular, interpretable structures such as communities, functional clusters, and multi-scale patterns, offering insights from network science into latent spatial dynamics.
\end{itemize}
\end{sloppypar}
\section{Related Work}
Spatio-temporal graph forecasting has been extensively studied in recent years, with most methods combining graph-based spatial modeling with recurrent or attention-based temporal encoders. Graph Convolutional Networks (GCNs) \cite{kipf2016semi} and Graph Attention Networks (GATs) have become the dominant paradigms for modeling spatial dependencies among nodes \cite{wu2021gnnsurvey, velivckovic2018gat}. These are typically integrated with temporal models such as Gated Recurrent Units (GRUs) \cite{chung2014empirical} or temporal attention \cite{xu2020spatial} to handle time-varying signals.

GAT-based forecasting models, such as ASTGCN \cite{guo2019attention}, STGAT \cite{9146162}, and STGSL \cite{zhang2021stgsl}, use attention to dynamically weigh node importance, but still operate under the constraint of a fixed adjacency matrix. 
However, these models typically assume a fixed or heuristically constructed adjacency matrix and compute attention scores at each forward pass, which can limit the model’s ability to capture long-range or evolving dependencies and reduce efficiency.

Recent works like STGSL \cite{zhang2021stgsl} and Graph WaveNet \cite{wu2019graphwavenet} propose learnable adjacency matrices while Neural Relational Inference (NRI) \cite{kipf2018neural} models latent dynamics via variational inference. Kazi et al. \cite{kazi2023differentiable} employs a differentiable graph module that learns soft graph structures using attention-based updates. Unlike these methods, we learn a single, persistent interaction matrix shared across time, enabling both efficiency and interpretability

Temporal modeling in spatio-temporal graphs is often handled via recurrent architectures such as GRUs or LSTMs, as seen in DCRNN \cite{li2018dcrnn} and GCN-GRU hybrids \cite{zhao2019t}. These models are effective in capturing sequential trends, but may under-use spatial flexibility when the graph structure is fixed. Transformer-based models such as ST-Transformer \cite{xu2020spatial} take a different approach by flattening spatio-temporal sequences and using self-attention across time and space. Although powerful, these models often ignore graph priors entirely. In contrast, we maintain a graph-based view while removing unnecessary constraints, striking a balance between structure-aware and structure-free attention.

Interpretable graph structures such as communities, functional clusters, and multi-scale patterns have long been central to network science, offering insight into the modular and hierarchical organization of complex systems \cite{bassett2017network, sales2007extracting}. In applied domains like traffic forecasting and brain networks, these structures often reflect latent functional roles rather than purely topological proximity \cite{zhang2021stgsl}. Methods such as hierarchical clustering \cite{sales2007extracting} and diffusion-based embeddings \cite{coifman2006diffusion} have been used to uncover community-aware patterns, but are often separate from learnable predictive modeling. 

Interpretability in graph learning is an active recent area of research, with methods like GNNExplainer \cite{ying2019gnnexplainer} aiming to extract human-interpretable substructures that influence model outputs. Other works like Deep Graph Infomax \cite{velickovic2020deep} emphasizes discovering clustered or community-aware representations. Data-driven representation learning \cite{hamilton2017representation, battaglia2018relational} is a growingly popular topic, but there is limited work on bringing insights from classical network analysis methods such as spectral analysis or community detection.

However, there exists little work in integrating conventional network analyses with data-driven learning. 
Our model learns node-node attention that results in the natural emergence of functional clusters, i.e.,groups of nodes that share predictive roles despite being physically distant, multi-scale organization (with different heads capturing different local vs global scales), and community structure. 
Through spectral and clustering analyses, we show that the learned topology supports topology-aware interpretability, bridging the predictive power of deep networks with the analytical strengths of graph and network science.

\section{Methodology}

Our proposed architecture, \textbf{InterGAT-GRU}, integrates a learnable interaction-based spatial encoder with a GRU-based temporal decoder for spatio-temporal forecasting. The core idea is to remove adjacency masking and conventional pairwise attention in Graph Attention Networks (GAT)~\cite{velivckovic2018gat}, and instead model spatial dependencies using a symmetric, fully learnable node-to-node interaction matrix. This allows the model to capture latent structural relationships without relying on predefined graph topology or dynamic attention computation.

\subsection{Problem Definition}

Spatio-temporal forecasting entails predicting future node-level signals over a graph-structured domain based on historical observations. This problem is defined over a graph \( G = (V, E) \) with \( |V| = N \) nodes and an adjacency matrix \( A \in \mathbb{R}^{N \times N} \). At each time step \( t \), nodes are associated with feature vectors in \( X_t \in \mathbb{R}^{N \times F} \). Given a sequence of past observations \( \{X_{t-n+1}, \dots, X_t\} \), the goal is to predict future node features \( \hat{X}_{t+1}, \dots, \hat{X}_{t+T} \in \mathbb{R}^{N \times F'} \).

\subsection{Interaction-Based Attention Modeling}

In conventional GAT-based models \cite{zhang2020spatial}, \cite{liu2020physical}, \cite{yu2023dgformer} attention scores are computed for neighboring nodes using a masked pairwise mechanism:
\begin{equation}
e_{ij} = 
\begin{cases}
\text{LeakyReLU} \left( \mathbf{a}^\top [\mathbf{W} \mathbf{x}_i \,\|\, \mathbf{W} \mathbf{x}_j] \right), & \text{if } A_{ij} > 0 \\
-\infty, & \text{otherwise}
\end{cases}
\label{eq:gat_attention}
\end{equation}
Where \(\mathbf{W}\) is a learnable weight matrix.
This enforces hard structural masking using \( A \) and limits interactions to predefined edges in the graph.
In contrast, we introduce a trainable, symmetric interaction matrix \( \mathbf{I} \in \mathbb{R}^{N \times N} \) that governs all pairwise dependencies, regardless of the original graph structure. This design allows any node to attend to any other, even when no edge exists in the original topology. We consider connected graphs. If the graph is disconnected, each connected component is treated separately.



\subsubsection{Feature Transformation}
Each node feature $x_i \in \mathbb{R}^F$ is projected into a latent space via a shared weight matrix $\mathbf{W} \in \mathbb{R}^{F \times F'}$:
\[
h_i = \mathbf{W} x_i.
\]
The input features are stacked to form the matrix $\mathbf{X} \in \mathbb{R}^{N \times F}$, and the corresponding transformed feature matrix $\mathbf{H} \in \mathbb{R}^{N \times F'}$ is given by:
\[
\mathbf{H} = \mathbf{X} \mathbf{W}
=
\begin{bmatrix}
h_1^\top,  
\hdots, 
h_N^\top
\end{bmatrix}^T,
\]

where $h_i \in \mathbb{R}^{F'}$ is the corresponding transformed feature of the input $x_i$, and $\mathbf{H} \in \mathbb{R}^{N \times F'}$ is the transformed feature matrix.


\subsubsection{Symmetric Interaction Encoding}
We process the interaction matrix as follows:
\begin{align}
\mathbf{I} &\leftarrow \frac{1}{2}(\mathbf{I} + \mathbf{I}^\top) \quad \text{(symmetrization)}. \\
\mathbf{I} &\leftarrow \text{LayerNorm}(\mathbf{I}) \quad \text{(stabilization)}. \\
\mathbf{I} &\leftarrow \text{softmax}(\mathbf{I}, \text{dim}=1) \quad \text{(row-wise normalization).}
\end{align}
\subsubsection{Spatial Aggregation}
\begin{figure}[!htbp]
  \centering
  \includegraphics[width=\linewidth]{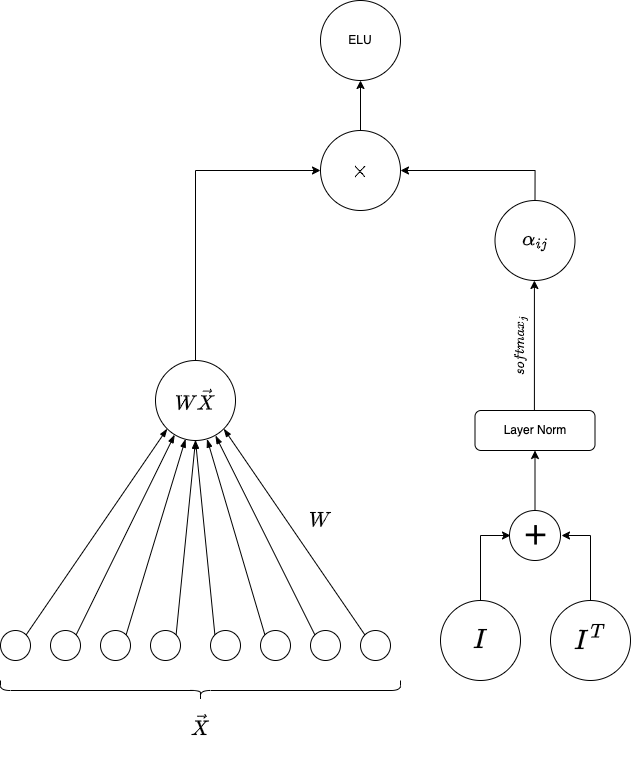}
  \caption{
InterGAT Architecture: Each input feature vector \( \vec{X} \) is transformed via a shared learnable weight matrix \( W \), producing \( W\vec{X} \). Attention coefficients \( \alpha_{ij} \) are computed using a normalized interaction matrix \( \mathbf{I} + \mathbf{I}^T \). The resulting attention weights modulate \( W\vec{X} \) through element-wise multiplication, followed by an ELU activation.}

  \Description{A woman and a girl in white dresses sit in an open car.}
  \label{fig: InterGATLayer}
\end{figure}

We then compute the spatially aggregated node embeddings \( \mathbf{Z} \in \mathbb{R}^{N \times F'} \) as:
\begin{equation}
\mathbf{Z}_i = \text{ELU}\left( \sum_{j=1}^{N} \mathbf{I}_{ij} \cdot \mathbf{h}_j \right),
\end{equation}
where ELU is Exponential Linear Unit activation function. The process is shown in Figure~\ref{fig: InterGATLayer}. The output \( \mathbf{Z} \) serves as the spatial representation at each time step and is passed to a GRU-based temporal module. 


\subsubsection{Key Differences from GAT}

Our design eliminates the need for pairwise attention score computation and adjacency masking, resulting in a structurally simplified and learnable attention mechanism:

\begin{itemize}
    \item \textbf{No masking:} Unlike GAT, which restricts attention via the adjacency matrix \( A \), our model allows all nodes to attend to each other by default.
    
    \item \textbf{No dynamic attention scores:} Traditional GAT computes \( e_{ij} \) based on learned projections and feature similarity. We replace this with a fixed interaction matrix \( \mathbf{I} \) shared across all time steps.
    
    \item \textbf{Fully learnable structure:} The interaction matrix \( \mathbf{I} \in \mathbb{R}^{N \times N} \) is trained end-to-end and regularized for symmetry and sparsity, capturing latent pairwise dependencies independent of temporal context.
\end{itemize}

\subsection{Regularization and Structural Constraints on the Interaction Matrix}

To ensure that the learned interaction matrix \( \mathbf{I} \in \mathbb{R}^{N \times N} \) captures meaningful and interpretable latent relationships between node pairs, we impose two key structural constraints:

\begin{itemize}
    \item \textbf{Sparsity Penalty:} To encourage a compact representation and suppress spurious or noisy interactions, we apply an \(\ell_1\) regularization term:
    \[
    \mathcal{L}_{\text{sparse}} = \lambda_{\text{sparse}} \cdot \|\mathbf{I}\|_1,
    \]
    where \( \lambda_{\text{sparse}} \) is a tunable hyperparameter controlling the strength of sparsity enforcement. This term promotes selective attention and enhances interpretability by pushing most values of \( \mathbf{I} \) toward zero.

    \item \textbf{Symmetry Constraint:} Since relational interactions should be bidirectional and symmetric in many spatial contexts, we explicitly enforce symmetry in the interaction matrix. During training, we replace the raw interaction matrix with its symmetrized counterpart:
    \[
    \mathbf{I} \leftarrow \frac{1}{2}(\mathbf{I} + \mathbf{I}^\top).
    \]
    This ensures \( I_{ij} = I_{ji} \) for all \( i, j \), and stabilizes learning while maintaining relational coherence across the graph structure.
\end{itemize}

The final loss used for training combines the main prediction loss with the regularization penalty:
\[
\mathcal{L}_{\text{total}} = \mathcal{L}_{\text{task}} + \mathcal{L}_{\text{sparse}},
\]
where \( \mathcal{L}_{\text{task}} \) is a forecasting loss that measures the discrepancy between the predicted node states and the ground truth at future time steps. Specifically, we use the Mean Squared Error (MSE) loss:
\[
\mathcal{L}_{\text{task}} = \frac{1}{BNT} \sum_{b=1}^{B} \sum_{t=1}^{T} \sum_{i=1}^{N} \| \hat{\mathbf{x}}^{(b)}_{i,t} - \mathbf{x}^{(b)}_{i,t} \|_2^2,
\]

Here, \( B \) denotes the batch size, \( T \) the number of future time steps being predicted, \( N \) the number of nodes, \( \hat{\mathbf{x}}^{(b)}_{i,t} \) the predicted feature vector for node \( i \) at time \( t \) in batch \( b \), and \( \mathbf{x}^{(b)}_{i,t} \) the corresponding ground truth.


\subsubsection{Complexity and Sparsity Control}
The standard GAT computes an $N \times N$ attention matrix, i.e it dynamically recomputes attention scores at each forward pass using input-dependent transformations (Eq.~\ref{eq:gat_attention}). In contrast, our model replaces these steps with a static and symmetric, learnable interaction matrix that is shared across time steps and trained end-to-end, reducing the overall per-epoch cost.

\paragraph{Parameter Count}
The learnable interaction matrix $\mathbf{I} \in \mathbb{R}^{N \times N}$ introduces $N^2$ parameters per attention head. In contrast, standard GAT layers require $O(F \cdot F')$ parameters for linear projections, plus additional parameters for attention vectors. 
Unlike prior work that addresses attention complexity via top-$k$ pruning or adjacency masking~\cite{shang2021gts}, our approach retains full pairwise connectivity and enforces soft sparsity using \(\ell_1\) regularization. This encourages the model to suppress weak or redundant edges while preserving the capacity to learn latent, long-range structural dependencies.

While both traditional GAT and InterGAT  exhibit quadratic scaling in theory, our approach eliminates input-dependent attention computation and sparse masking resulting in a leaner forward-backward graph and improved runtime efficiency, particularly on small-to-medium graphs like SZ-Taxi ($N=156$) and Los-loop ($N=207$). For small graphs (e.g., $N < 300$), the cost of storing $\mathbf{I}$ remains modest and comparable to typical projection layers. However, we note that $\mathbf{I}$ scales quadratically with the number of nodes, making parameter count and potential overfitting a concern for large graphs. To address this, one may use low-rank approximation techniques such as LoRA~\cite{hu2021lora}, which would reparameterize $\mathbf{I}$ as the product of two low-rank matrices, significantly reducing both parameter count and computational costs, to mitigate this.

\subsection{Temporal Modeling with GRU}

After extracting spatial representations from each time step using the Dense Graph Attention Layer, we model temporal dependencies across these embeddings using a Gated Recurrent Unit (GRU)~\cite{cho2014gru}. GRUs are a type of recurrent neural network designed to effectively capture sequential patterns while mitigating the vanishing gradient problem that arises in traditional RNNs.

Let $\{\mathbf{H}_{t-n+1}, \mathbf{H}_{t-n+2}, \dots, \mathbf{H}_t\}$ denote the sequence of spatial embeddings produced by the InterGAT layer over the last $n$ time steps, where each $\mathbf{H}_\tau \in \mathbb{R}^{N \times F'}$ represents the hidden representation of all $N$ nodes at time $\tau$. These embeddings are fed into the GRU in temporal order.

The GRU updates its hidden state for each node independently at each time step using the following equations:
\begin{align}
\mathbf{z}_t &= \sigma\left(\mathbf{W}_z \mathbf{H}_t + \mathbf{U}_z \mathbf{h}_{t-1}\right), \\
\mathbf{r}_t &= \sigma\left(\mathbf{W}_r \mathbf{H}_t + \mathbf{U}_r \mathbf{h}_{t-1}\right), \\
\tilde{\mathbf{h}}_t &= \tanh\left(\mathbf{W}_h \mathbf{H}_t + \mathbf{U}_h (\mathbf{r}_t \odot \mathbf{h}_{t-1})\right), \\
\mathbf{h}_t &= (1 - \mathbf{z}_t) \odot \mathbf{h}_{t-1} + \mathbf{z}_t \odot \tilde{\mathbf{h}}_t,
\end{align}

Here, $\mathbf{z}_t$ and $\mathbf{r}_t$ are the update and reset gates, respectively; $\mathbf{h}_t \in \mathbb{R}^{N \times F'}$ is the hidden state at time $t$, and $\odot$ denotes element-wise multiplication. The weights $W_*$ and $U_*$ are learned during training.

The final hidden state $\mathbf{h}_t$ captures the temporal dynamics aggregated over the historical sequence and is passed to a fully connected linear decoder to generate the prediction for each node:
\begin{equation}
\hat{\mathbf{X}}_{t+\tau} = \text{Linear}(\mathbf{h}_t), \quad \tau \in \{1, \dots, T\}.
\end{equation}

For multi-step forecasting, we adopt a sequence-to-sequence decoder strategy: at each future step, the GRU continues to generate hidden states using either the previous prediction or a scheduled teacher-forced ground truth input, depending on the training mode. 
This allows the model to make accurate, temporally consistent predictions over the entire forecast horizon.








\section{Experiments}

We evaluate the proposed \textbf{InterGAT-GRU} architecture on two benchmark traffic datasets to assess its effectiveness in spatio-temporal forecasting. The InterGAT spatial encoder consists of a single graph attention layer with 4 attention heads that each learn an independent interaction matrix and have an output dimension of 32. The temporal dependencies are modeled using a single-layer GRU with 128 hidden units. Predictions are made  15, 30, 45, 60 minutes into the future for SZ-Taxi and 15, 30, 45, and 60 minutes ahead for Los-Loop Dataset. The model is trained using a single NVIDIA A100 40GB GPU with CUDA 11.8 and PyTorch 2.0. for a max of 100 epochs.  


\subsection{Data Description}
We evaluate our model using two real-world benchmark traffic datasets, both of which contain spatial and temporal traffic speed data. In both datasets, traffic speed serves as the target variable, and each sample 
consists of (i) a graph structure, represented by an adjacency matrix that captures spatial connectivity between road segments or sensors, and (ii) temporal features, which are historical traffic speed readings over time at each node. The data is normalized to the range [0, 1], and we use 80\% of the data for training and 20\% for testing.

\begin{itemize}
    \item \textbf{SZ-Taxi} : This dataset contains taxi trajectory data collected in Shenzhen, China, from January 1 to January 31, 2015. We focus on 156 major roads in the Luohu District. The dataset consists of two components: a $156 \times 156$ adjacency matrix representing spatial road connectivity, and a traffic speed matrix where each row corresponds to a road and each column records the average speed aggregated every 15 minutes.

    \item \textbf{Los-Loop}: Collected via loop detectors on the highways of Los Angeles County, this dataset includes traffic speed data from 207 sensors recorded between March 1 and March 7, 2012. The adjacency matrix is constructed based on the pairwise distances between sensors in the traffic network. Traffic speed is aggregated every 5 minutes. Missing values are handled via linear interpolation.
\end{itemize}

\subsection{Model Implementation Details}
The proposed InterGAT-GRU model is implemented using PyTorch. For training, we use the Mean Squared Error (MSE) as the loss function to minimize the difference between predicted and actual traffic speed values. The optimizer used is Adam with a batch size of 32, a default learning rate of $1 \times 10^{-3}$ and weight decay set to $1 \times 10^{-5}$ for regularization.

The InterGAT spatial encoder consists of a single graph attention layer with \textbf{4 attention heads}. Each head uses its own independent interaction matrix and attention parameters. The output of each head has a dimensionality of 32, resulting in a final concatenated spatial embedding of size 128 per node. The temporal dependencies are modeled using a single-layer GRU with 128 hidden units (to match the spatial output dimension).

The model is trained for a maximum of 100 epochs, with early stopping based on validation MAE. A mini-batch size of 32 is used during training. Each input sequence consists of $n=12$ historical time steps. For the SZ-Taxi dataset, each step corresponds to a 15-minute time interval, and the model is evaluated on prediction horizons of 1, 2, 3, and 4 steps (i.e., 15, 30, 45, and 60 minutes into the future). For the Los-loop dataset, which uses 5-minute intervals, the model is evaluated on horizons of 3, 5, 9, and 12 steps (i.e., 15, 30, 45, and 60 minutes ahead). Dropout with a rate of 0.3 is applied after the InterGAT layer to prevent overfitting.

All experiments are conducted on a single NVIDIA A100 GPU. The model typically converges within 20 to 40 epochs, depending on the dataset and the prediction horizon (Figure~\ref{fig:forecast_loss_comparison} {see Appendix \ref{appendix:training}}).

\section{Results}
\subsection{Performance on traffic datasets}

To isolate the impact of our architectural modifications, we compare InterGAT-GRU against a baseline GAT-GRU model under identical training conditions. Results are averaged over five independent runs with different random seeds to ensure robustness.
As shown in Table~\ref{tab:gat_vs_InterGAT}, InterGAT-GRU consistently achieves lower Mean Absolute Error (MAE) and higher accuracy (measured via Frobenius norm see Appendix \ref{appendix:evaluation}) across all horizons on both datasets, demonstrating its effectiveness in modeling latent spatial dependencies without relying on explicit neighborhood structures or attention scores.
Notably, the largest gains are observed at the 60-minute horizon, where MAE drops from \(4.67 \pm 0.08\) to \(2.97 \pm 0.05\) on SZ-Taxi, and from \(7.35 \pm 0.10\) to \(4.82 \pm 0.07\) on Los-Loop. This suggests that the learned global structure better captures long-range dependencies, that are important for predictions of extended time horizons.

\begin{table*}[ht]
\centering
\caption{Performance comparison between GAT-GRU and InterGAT-GRU on SZ-Taxi and Los-Loop datasets across multiple horizons. Results are reported as mean ± standard deviation over five independent runs with random initializations. Evaluation metrics are Mean Average Error (MAE) and Accuracy defined in terms of Frobenius norm between predicted and actual valus (see Appendix \ref{appendix:evaluation} for definitions)}.
\resizebox{\textwidth}{!}{%
\begin{tabular}{cc|cc|cc||cc|cc|cc}
\toprule
\multicolumn{6}{c||}{\textbf{SZ-Taxi}} & \multicolumn{6}{c}{\textbf{Los-Loop}} \\
\cmidrule{1-12}
\textbf{Horizon} & \textbf{Model} & \textbf{MAE} & \textbf{Accuracy} & ~ & ~ &
\textbf{Horizon} & \textbf{Model} & \textbf{MAE} & \textbf{Accuracy} & ~ & ~ \\
\midrule
\multirow{2}{*}{15 min} & GAT-GRU        & 4.6082 ± 0.061 & 0.5758 ± 0.013 & ~ & ~ &
\multirow{2}{*}{15 min} & GAT-GRU        & 7.0257 ± 0.093 & 0.8298 ± 0.012 & ~ & ~ \\
                        & InterGAT-GRU   & 2.8558 ± 0.038 & 0.7071 ± 0.008 & ~ & ~ &
                        & InterGAT-GRU   & 3.7752 ± 0.057 & 0.8953 ± 0.007 & ~ & ~ \\
\multirow{2}{*}{30 min} & GAT-GRU        & 4.6391 ± 0.067 & 0.5733 ± 0.014 & ~ & ~ &
\multirow{2}{*}{30 min} & GAT-GRU        & 7.1485 ± 0.087 & 0.8256 ± 0.011 & ~ & ~ \\
                        & InterGAT-GRU   & 2.9015 ± 0.041 & 0.7041 ± 0.009 & ~ & ~ &
                        & InterGAT-GRU   & 4.0846 ± 0.062 & 0.8873 ± 0.008 & ~ & ~ \\
\multirow{2}{*}{45 min} & GAT-GRU        & 4.6581 ± 0.072 & 0.5718 ± 0.014 & ~ & ~ &
\multirow{2}{*}{45 min} & GAT-GRU        & 7.3383 ± 0.095 & 0.8211 ± 0.012 & ~ & ~ \\
                        & InterGAT-GRU   & 2.9210 ± 0.049 & 0.7028 ± 0.010 & ~ & ~ &
                        & InterGAT-GRU   & 4.3882 ± 0.066 & 0.8768 ± 0.009 & ~ & ~ \\
\multirow{2}{*}{60 min} & GAT-GRU        & 4.6701 ± 0.076 & 0.5712 ± 0.016 & ~ & ~ &
\multirow{2}{*}{60 min} & GAT-GRU        & 7.3485 ± 0.098 & 0.8163 ± 0.014 & ~ & ~ \\
                        & InterGAT-GRU   & 2.9652 ± 0.053 & 0.6985 ± 0.011 & ~ & ~ &
                        & InterGAT-GRU   & 4.8219 ± 0.070 & 0.8639 ± 0.010 & ~ & ~ \\
\bottomrule
\end{tabular}
}
\label{tab:gat_vs_InterGAT}
\end{table*}


Accuracy gains follow a similar pattern, with InterGAT-GRU achieving improvements of up to 13\% on short horizons and maintaining strong performance across all settings. Furthermore, the low standard deviation across runs demonstrates that InterGAT-GRU not only achieves better average performance but also exhibits greater training stability. Taken together, these results confirm that replacing masked attention with a dense, learnable interaction matrix offers a more expressive and robust inductive bias for spatio-temporal forecasting, particularly when adjacency structure is sparse or long-range interactions are crucial.


\subsection{Interaction Matrix Sparsity}
\begin{figure}[!htbp]
    \centering
    \includegraphics[width=\linewidth]{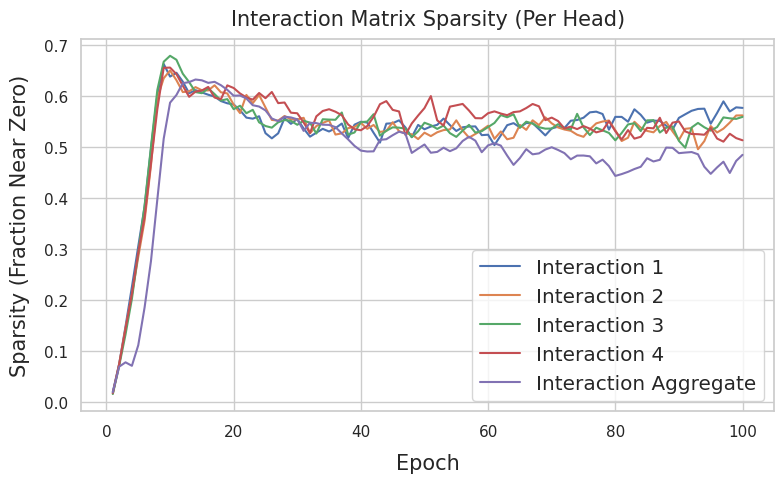}
    \caption{Sparsity evolution of \( \mathbf{I} \) per head, defined as fraction of values \( < 10^{-4} \). The model promotes selective attention across all heads.}
    \label{fig:sparsity_plot}
    \Description{A line plot showing the performance increase over time.}
\end{figure}

To quantify the learned structure \( \mathbf{I} \), we compute its element sparsity, defined as the proportion of values close to zero (i.e., \( |\mathbf{I}_{ij}| < 10^{-4} \)). As shown in Figure~\ref{fig:sparsity_plot}, sparsity increases sharply during early training, stabilizing between 50\% and 70\% across attention heads.

This early increase in sparsity suggests that the model rapidly learns hidden latent structure from the data and suppresses weak, noisy, or redundant interactions. 
Notably, this sparsity emerges naturally from the soft $\ell_1$ regularization without explicit masking. 

In parallel, we observe that the Frobenius norm of \( \mathbf{I} \) also converges early and stabilizes across training Figure~\ref{fig:interaction_norm} (see Appendix \ref{appendix:convergence}), indicating that the magnitude and distribution of interactions have settled into a stable structure.

\textit{Visualizing Interaction matrices:} We visualize the \emph{interaction matrices} produced by each attention head. 
Each head is min-max normalized (values are scaled to  $[0,1]$ ), and only the top 2\% of values are visualized to reveal the most salient attention-driven connections (see Figure~\ref{fig:interaction_heatmaps} in Appendix \ref{appendix: int_matrix}).


\subsection{Training Efficiency Analysis}
\label{sec:training_time}

We compare the training efficiency of InterGAT-GRU against the standard BaseGAT-GRU on the SZ-Taxi and Los loop dataset across all prediction horizons. As shown in Table~\ref{tab:intergat_runtime}, InterGAT-GRU reduces the training time by 60\% for the SZ-Taxi dataset and 70\% for the Los-Loop dataset across all time horizons, in comparison with an identical baseline GAT-GRU. Both models are trained for 100 epochs on the same forecasting horizon task, using identical batch sizes, optimizer settings, and hardware configurations.

This empirical efficiency gain stems from the architectural simplification in InterGAT. By replacing dynamic, edge-conditioned attention computations with a static, learnable interaction matrix \( \mathbf{I} \), we eliminate per-node attention score calculations and row-wise softmax normalization, which dominate the computational cost in GATs. Notably, the forward pass in InterGAT is significantly faster. 


Note that absolute performance may vary across hardware setups or larger graphs. Nevertheless, these results suggest that InterGAT provides a leaner and more efficient training pipeline in practice, making it appealing for time-sensitive or resource-constrained spatio-temporal forecasting scenarios.

\begin{table*}[ht]
\centering
\caption{Training efficiency metrics grouped by dataset and forecast horizon. All values are averaged over 100 epochs.}
\begin{tabular}{lclccc}
\toprule
\textbf{Dataset} & \textbf{Forecast Horizon} & \textbf{Model} & \textbf{Epoch Time (s)} & \textbf{Total Train Time (min)} & \textbf{Forward / Backward Pass (s)} \\
\midrule

\multirow{8}{*}{SZ-Taxi} 
  & \multirow{2}{*}{15min}
    & BaseGAT-GRU   & 5.468 & 9.114 & 2.676 / 2.666 \\
  &                & InterGAT-GRU  & 2.228 & 3.714 & 0.857 / 1.267 \\[0.3em]
  & \multirow{2}{*}{30min}
    & BaseGAT-GRU   & 5.473 & 9.121 & 2.680 / 2.670 \\
  &                & InterGAT-GRU  & 2.231 & 3.720 & 0.852 / 1.269 \\[0.3em]
  & \multirow{2}{*}{45min}
    & BaseGAT-GRU   & 5.479 & 9.128 & 2.699 / 2.676 \\
  &                & InterGAT-GRU  & 2.235 & 3.726 & 0.850 / 1.274 \\[0.3em]
  & \multirow{2}{*}{60min}
    & BaseGAT-GRU   & 5.484 & 9.132 & 2.702 / 2.682 \\
  &                & InterGAT-GRU  & 2.237 & 3.729 & 0.848 / 1.282 \\

\midrule

\multirow{8}{*}{Los-Loop} 
  & \multirow{2}{*}{15min}
    & BaseGAT-GRU   & 6.067 & 10.618 & 3.036 / 2.927 \\
  &                & InterGAT-GRU  & 1.588 & 2.912 & 0.6091 / 0.8904 \\[0.3em]
  & \multirow{2}{*}{30min}
    & BaseGAT-GRU   & 6.073 & 10.792 & 3.040 / 2.930 \\
  &                & InterGAT-GRU  & 1.593 & 2.954 & 0.6103 / 0.8923 \\[0.3em]
  & \multirow{2}{*}{45min}
    & BaseGAT-GRU   & 6.090 & 10.843 & 3.048 / 2.951 \\
  &                & InterGAT-GRU  & 1.597 & 3.016 & 0.6109 / 0.8957 \\[0.3em]
  & \multirow{2}{*}{60min}
    & BaseGAT-GRU   & 6.104 & 10.915 & 3.053 / 2.981 \\
  &                & InterGAT-GRU  & 1.603 & 3.107 & 0.6112 / 0.900 \\
\bottomrule
\end{tabular}
\label{tab:intergat_runtime}
\end{table*}



\subsection{Ablation Study on Interaction Matrix Formulations}

We conducted an extensive ablation study comparing alternate formulations of the interaction matrix \( \mathbf{I'} \) in InterGAT. Our goal was to contrast our symmetric and fully learnable interaction matrix\( \mathbf{I} \) with other conventional formalisms such as attention \cite{velivckovic2018gat} or fixed adjacency-based mechanisms \cite{kipf2016semi}. 
We explored different parameterizations of the matrix \( \mathbf{I'} \in \mathbb{R}^{N \times N} \), each representing a different inductive bias or level of learnability:
\begin{equation}
\mathbf{I'} =
\begin{cases}
0, & \text{(No bias — standard dense attention)} \\
\mathbf{I}, & \text{(Fully trainable matrix, symmetrized)} \\
\mathbf{A}, & \text{(Binary adjacency matrix)} \\
\mathbf{W} \cdot \mathbf{A}, & \text{(Trainable weighted adjacency)} \\
\mathbf{W} \cdot \mathbf{C}, & \text{(Trainable weighted covariance)} \\
\tilde{\mathbf{A}}, & \text{(Spectral clustering-based clustered adjacency)}
\end{cases}
\label{eq:I_variants}
\end{equation}
\noindent
Here, \( A \in \{0, 1\}^{N \times N} \) is the original binary adjacency matrix, \( \mathbf{W} \in \mathbb{R}^{N \times N} \) is a fully trainable matrix, \( C \in \mathbb{R}^{N \times N} \) represents the empirical pairwise covariance of node characteristics \cite{cavallo2024spatiotemporal}, and \( \tilde{A} \) is a block-diagonal binary adjacency matrix (see Appendix~\ref{appendix:spectral}) obtained via spectral clustering \cite{ng2002spectral} with a cluster size of 10.

\begin{table}[ht]
\centering
\caption{Ablation study using different forms of the interaction matrix \( \mathbf{I'} \). We report Accuracy in terms of Frobenius norm.}
\begin{tabular}{l|cc}
\toprule
\textbf{Interaction Matrix \( \mathbf{I'} \)} & \textbf{MAE} & \textbf{Accuracy} \\
\midrule
Base GAT (\( \mathbf{I'} = 0 \))                          & 4.6082  & 0.5758  \\
Adjacency matrix (\( \mathbf{I'} = \mathbf{A} \))              & 4.5901  & 0.5766  \\
Weighted adjacency (\( \mathbf{I'} = \mathbf{W} \cdot \mathbf{A} \))    & 3.0614  & 0.6925  \\
Weighted covariance (\( \mathbf{I'} = \mathbf{W} \cdot \mathbf{C} \))   & 3.0146  & 0.6897  \\
Spectral Similarity Matrix (\( \mathbf{I'} = \tilde{\mathbf{A}} \)) & 4.6250  & 0.5742  \\
Learnable, symmetrized \( \mathbf{I'} = \tilde{\mathbf{I}} \))                    & \textbf{2.8558}  & \textbf{0.7071}  \\
\bottomrule
\end{tabular}
\label{tab:ablation_masking_single}
\end{table}


The learnable matrix outperforms all alternatives as can be seen from Table \ref{tab:ablation_masking_single}, with lower MAE and higher accuracy, affirming its expressiveness and flexibility. 

\subsection{Spectral Analysis of the learned Interaction Matrix}
\label{sec:spectral_analysis}
To better understand the structural properties encoded by the learned interaction matrix, we analyze its eigenvectors in terms of smoothness and localization—two key characteristics in graph signal processing.

\begin{figure}[ht]
    \centering
    \includegraphics[width=0.9\linewidth]{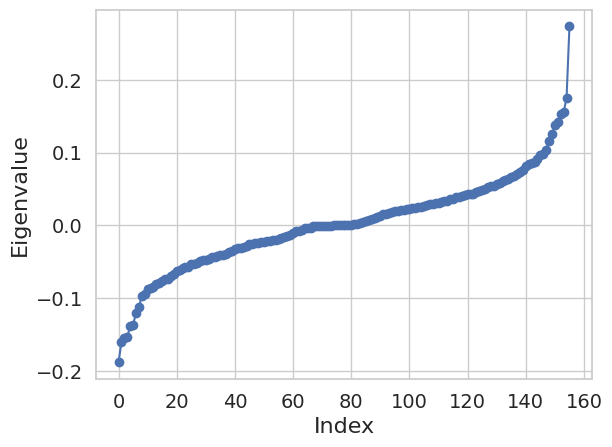}
    \caption{Eigenvalue spectrum of the learned interaction matrix \( \mathbf{I} \) in the final GAT layer.}
    \Description{Eigenvalue spectrum of the learned interaction matrix \( \mathbf{I} \) in the final GAT layer.}
    \label{fig:eigenvalue_spectrum}
\end{figure}

Figure~\ref{fig:eigenvalue_spectrum} illustrates the eigenvalue spectrum of the learned interaction matrix \( \mathbf{I} \) in the final GAT layer. The eigenvalues span a range from approximately \(-0.2\) to \(+0.25\), with most eigenvalues being nonzero indicating that \( \mathbf{I} \) is near full rank with a rank of 155 for the $\mathbf{I}$ corresponding to the SZ-Taxi dataset. This stands in contrast to traditional GATs\cite{velivckovic2018gat}, where over-smoothed or poorly regularized attention weights can lead to rank-deficient interaction structures. 


\subsubsection{Smoothness via Dirichlet Energy}
We define the \textit{Dirichlet energy} of an eigenvector $\mathbf{v} \in \mathbb{R}^N$ with respect to the graph structure encoded by \( \mathbf{I} \) as:
\[
\mathcal{E}(\mathbf{v}) = \mathbf{v}^\top \tilde{\mathbf{L}} \mathbf{v} = \sum_{i,j} \mathbf{I}_{ij} (v_i - v_j)^2,
\]
where $\tilde{\mathbf{L}} = \mathbf{D} - \mathbf{I}$, $\mathbf{D}_{ii} = \sum_j \mathbf{I}_{ij}$, and $\mathbf{I}_{i,j}$ is the scalar component of $\mathbf{I}$.

\paragraph{Spectral Interpretation.}
Unlike classical graph Laplacians derived from adjacency matrices, our Laplacian $\tilde{\mathbf{L}} = \mathbf{D} - \mathbf{I}$ is built from a \textit{learned attention matrix} $\mathbf{I}$. 
Because the interaction matrix $\mathbf{I}$ is a learned attention mechanism, not a fixed adjacency matrix, its spectrum reflects which node relationships are most useful for task-specific prediction, rather than which nodes are physically connected. Note that $\mathbf{I}$ can contain negative values, and the corresponding $\tilde{\mathbf{L}}$ is not necessarily positive semi-definite. As a result, the spectral structure of $\tilde{\mathbf{L}}$ diverges from diffusion-based graphs. 

\begin{figure}[ht]
    \centering
    \includegraphics[width=\linewidth]{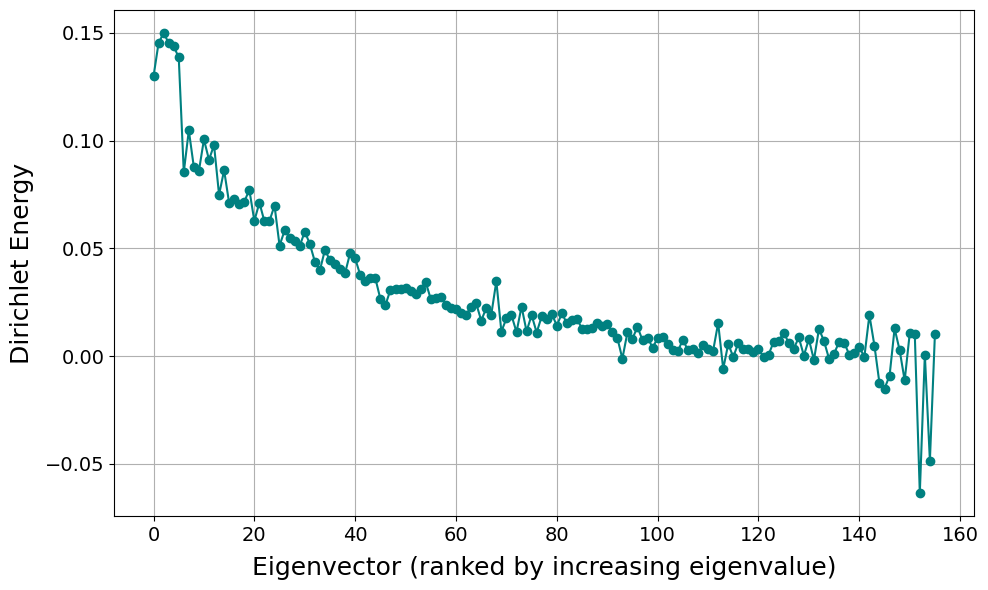}
\caption{
Dirichlet Energy of the eigenvectors of the learned interaction matrix \( \mathbf{I} \), sorted by increasing eigenvalue index.}
\Description{Dirichlet Energy of the eigenvectors of the learned interaction matrix \( \mathbf{I} \), sorted by increasing eigenvalue index.}
    \label{fig:dirichlet}
\end{figure}

We observe, in Figure \ref{fig:dirichlet}, that \textbf{low-index eigenvectors (small eigenvalues)} exhibit \textbf{high Dirichlet energy}, indicating that the most predictive modes emphasize sharp contrasts across key node pairs, rather than smooth propagation over the spatial structure. 

\subsubsection{Localization via Inverse Participation Ratio}
To assess how localized an eigenvector is, i.e., how concentrated its energy is on a small subset of nodes, we compute the \textit{Inverse Participation Ratio (IPR)}:\[
\mathrm{IPR}(\mathbf{v}) = \sum_{i=1}^N v_i^4,
\]
for a normalized eigenvector with $\|\mathbf{v}\|_2 = 1$. The IPR is close to $1/N$ for a uniformly spread vector and approaches 1 when concentrated on a single node. Thus, higher IPR values indicate stronger localization. Note that since $\mathbf{I}$ doesn't capture structural connectivity, the term `localized' refers to a small subset of most discriminative nodes that the model has learned to focus attention on. These nodes are not necessarily neighbors in space or topology.

\begin{figure}[!htbp]
    \centering
    \includegraphics[width=\linewidth]{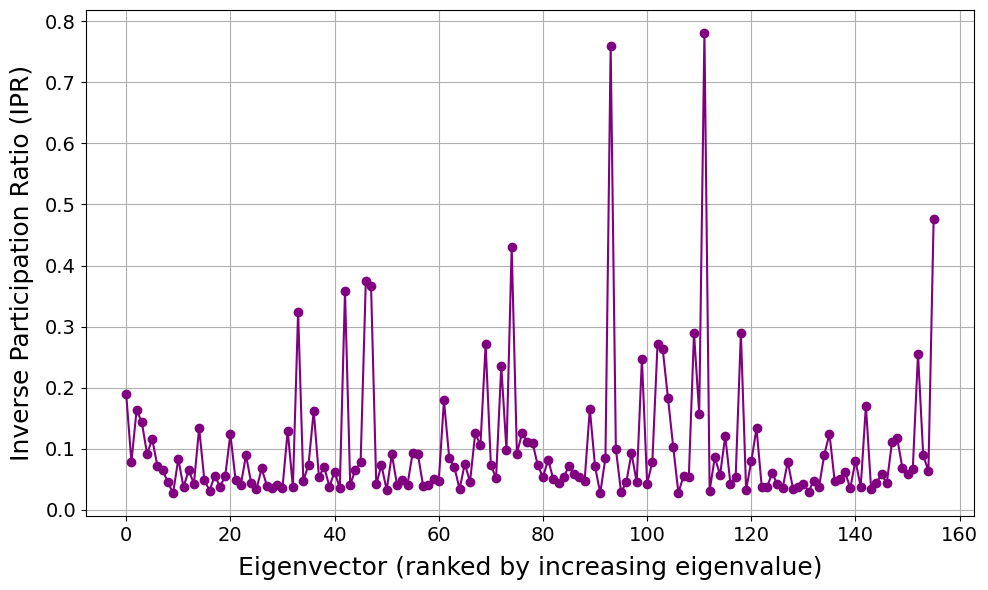}
    \caption{Inverse Participation Ratio (IPR) of eigenvectors of the learned interaction matrix \( \mathbf{I} \), plotted against eigenvalue index (sorted in ascending order).}
    \Description{Inverse Participation Ratio (IPR) of eigenvectors of the learned interaction matrix \( \mathbf{I} \), plotted against eigenvalue index (sorted in ascending order).}
    \label{fig:ipr}
\end{figure}

As seen in Figure \ref{fig:ipr}, a few eigenvectors in the middle of the spectrum have high IPR, i.e., these modes capture sparse, high-contrast interactions reflecting localized node-specific patterns focused on a small set of key nodes. In contrast, many eigenvectors at both ends of the spectrum have low IPR, indicating broad diffused attention patterns that encode diffusive global signals spanning large graph regions. Thus, the use multiple heads in our model promotes \textbf{multi-scale pattern discovery} capture of a mix of global and localized attention patterns, enabling it to represent both broad trends and sharp, node-specific distinctions in the data.

\begin{figure}[!htbp]
    \centering
    \includegraphics[width=0.75\linewidth]{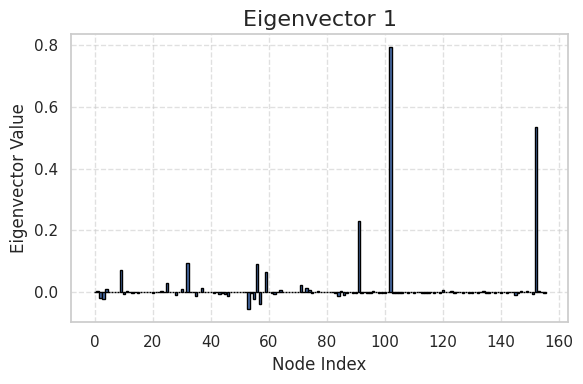} \\
    \includegraphics[width=0.75\linewidth]{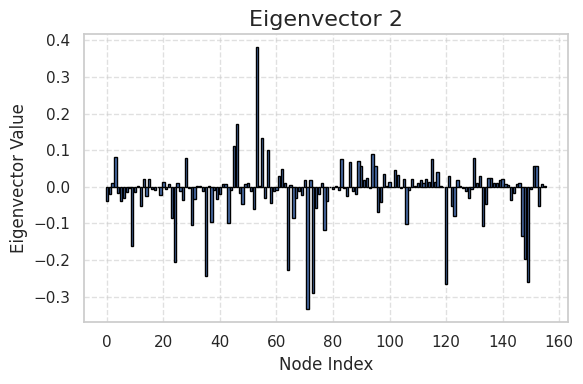} \\
    \includegraphics[width=0.75\linewidth]{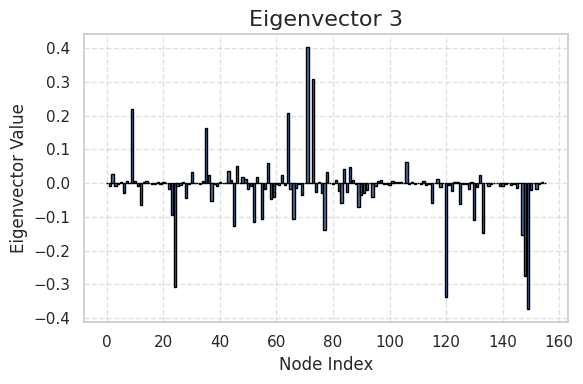}
    \caption{Top three eigenvectors (156--154) of the final layer interaction matrix \( \mathbf{I} \). Each eigenvector highlights different frequency components over the graph structure.}
    \Description{Top three eigenvectors (156--154) of the final layer interaction matrix \( \mathbf{I} \). Each eigenvector highlights different frequency components over the graph structure.}
    \label{fig:eigenvectors}
\end{figure}

To demonstrate this further, we examine the top three eigenvectors of the final-layer interaction matrix $\mathbf{I}$. As shown in Figure~\ref{fig:eigenvectors}, the largest eigenvector is highly localized, with a single node dominating the spectral mode, capturing concentrated attention on key spatial regions - effectively enabling node-specific signal amplification, whereas the second and third eigenvector captures global relationships.

\subsection{Topology-Aware Interpretability via Community Detection}
To investigate how the learned attention matrix \( \mathbf{I} \) relates to graph structure, we analyze its alignment with community partitions.  
Specifically, we perform spectral clustering~\cite{ng2002spectral} on the static adjacency matrix to obtain community partitions and assess whether \( \mathbf{I} \) assigns stronger attention weights within communities than across them.

Given a range of cluster counts \( k \), we partition the nodes into \( \{\mathcal{C}_1, \ldots, \mathcal{C}_k\} \) using the graph Laplacian derived from the static adjacency matrix. For each clustering configuration, we compute:
\begin{itemize}
    \item \textbf{Intra-community mean} \( \mu_{\text{intra}} \): average learned interaction strength between nodes within the same cluster
    \item \textbf{Inter-community mean} \( \mu_{\text{inter}} \): average learned interaction strength between nodes across different clusters
\end{itemize}

This analysis is repeated for each attention head in InterGAT, and Figure~\ref{fig:spectral_all_heads} shows the results across $k$ (see Appendix \ref{appendix:community}). To quantify the alignment between the learned interaction matrix \( \mathbf{I} \) and spectral communities, we compute the \textbf{normalized contrast metric}:

\[
\text{Contrast}_k = \frac{\mu_{\text{intra},k} - \mu_{\text{inter},k}}{\mu_{\text{inter},k} + \epsilon},
\]

where \( \mu_{\text{intra},k} \) and \( \mu_{\text{inter},k} \) denote the average intra- and inter-community interaction values for each cluster size \( k \in \{2, 3, \ldots, 31, 32\} \), spanning 30 cluster configurations, and \( \epsilon = 10^{-6} \) is a small constant added for numerical stability.

We also compute the corresponding standard deviation via error propagation to assess consistency across cluster sizes.

\[
\text{Std}_k = \frac{\sqrt{ \sigma_{\text{intra},k}^2 + \sigma_{\text{inter},k}^2 }}{ \mu_{\text{inter},k} + \epsilon },
\]

where \( \sigma_{\text{intra},k} \) and \( \sigma_{\text{inter},k} \) are the standard deviations of the intra- and inter-community interactions, respectively.



As summarized in Table~\ref{tab:community_contrast}, several attention heads, especially Head 4, exhibit strong and consistent contrast between intra- and inter-community interactions, suggesting that the model \textbf{learns community-aware attention patterns}. Other heads, such as Heads 2 and 3, show lower contrast and higher variability, indicating they may focus on more diffuse or global relationships rather than modular structure. 
The learned interaction matrix identifies \textbf{functional clusters} by assigning consistently higher attention weights within latent node groups that share predictive roles, even when they are not topologically adjacent

These results reinforce trends observed in Figure~\ref{fig:spectral_all_heads}: different attention heads specialize in capturing different types of relational patterns, some focus on local cohesion within communities, while others model global or cross-community dependencies. This division of labor enhances the model's expressive capacity and interpretability.


\begin{table}[ht]
\centering
\caption{Normalized community contrast for each attention head, averaged across cluster sizes ($k$) between 2 and 32. Higher mean values indicate stronger alignment with community structure, and lower standard deviation reflects consistent selectivity across $k$.}
\begin{tabular}{lcc}
\toprule
\textbf{Interaction Head} & \textbf{Mean Contrast} & \textbf{Contrast Std. Dev.} \\
\midrule
Interaction 1 & 1.491753 & 0.132830 \\
Interaction 2 & 0.757297 & 0.090759 \\
Interaction 3 & 0.637930 & 0.121607 \\
Interaction 4 & 3.270199 & 0.341248 \\
Interaction Agg. & 1.473864 & 0.280454 \\
\bottomrule
\end{tabular}
\label{tab:community_contrast}
\end{table}

\section{Conclusion \& Future Work}

We introduced InterGAT-GRU, a simplified spatio-temporal forecasting model that replaces adjacency masking and dynamic attention with an interpretable and learnable interaction matrix. This design improves efficiency and outperforms standard GAT-GRU baselines on benchmark traffic datasets across multiple time horizons, while requiring less than half the training time of the baseline. 

Spectral and community-based analyses show that the learned matrix reveals meaningful structure both localized and global modes, demonstrating interpretable features such as community-awareness, functional clusters, and capturing multi-scale patterns ranging from localized to global modes, enabling the model to recover latent graph topology without supervision. 
These patterns are not only predictive but also interpretable. This enables a form of topology-aware interpretability bridging deep learning with conventional graph-based methods for improved prediction and explainability of spatio-temporal graph-based learning.

In future work, we aim to extend this framework to dynamic graphs and explore low-rank or sparse formulations to support large-scale or real-time deployments. Our implementation can be found here \href{https://anonymous.4open.science/r/Beyond-Attention-ED5D/}{\color{blue}https://anonymous.4open.science/r/Beyond-Attention-ED5D/}.

\bibliographystyle{ACM-Reference-Format}
\bibliography{main}


\appendix
\section{Appendix}

\subsection{Spectral Clustering-Based Adjacency Rewiring}
\label{appendix:spectral}

To investigate the effect of explicit community structure on forecasting performance, we include a spectral clustering-based variant \( \tilde{A} \) in our ablation. Given the original adjacency matrix \( A \in \mathbb{R}^{N \times N} \), we apply spectral clustering to the graph Laplacian to identify \( k \) communities \( \mathcal{C}_1, \ldots, \mathcal{C}_k \). The clustered adjacency matrix is then defined as:
\begin{equation}
\tilde{A}_{ij} =
\begin{cases}
1 & \text{if } i \neq j \text{ and } i, j \in \mathcal{C}_\ell \text{ for some } \ell \in \{1, \ldots, k\} \\
0 & \text{otherwise}
\end{cases}
\label{eq:clustered_adj}
\end{equation}

This block-structured adjacency enforces full intra-cluster connectivity and eliminates inter-cluster edges, thereby encoding a strong inductive bias aligned with latent community structure. When used as \( \mathbf{I} \), this variant effectively replaces the learned interaction matrix with a hard-coded structural prior.



\subsection{Interaction Matrix Convergence}
\label{appendix:convergence}
\begin{figure}[ht]
    \centering
    \includegraphics[width=\linewidth]{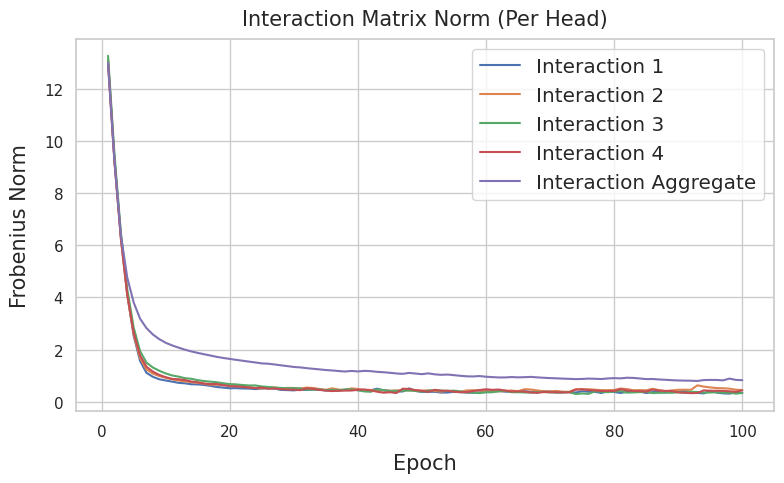}
    \caption{Frobenius norm of the interaction matrix \( \mathbf{I} \) per attention head across training epochs. Most heads converge to stable, low-norm states.}
    \Description{Frobenius norm of the interaction matrix \( \mathbf{I} \) per attention head across training epochs. Most heads converge to stable, low-norm states.}
    \label{fig:interaction_norm}
\end{figure}

In Figure~\ref{fig:interaction_norm}, we track the Frobenius norm of the interaction matrices  \( \mathbf{I} \) learned by each attention head over training epochs to assess their convergence.

\subsection{Commmunity-Aware Interaction Matrix Analysis}
\label{appendix:community}

Given a range of cluster counts \( k \), we partition the nodes into \( \{\mathcal{C}_1, \ldots, \mathcal{C}_k\} \) using the graph Laplacian derived from the static adjacency matrix. For each clustering configuration, we compute:
\begin{itemize}
    \item \textbf{Intra-community mean} \( \mu_{\text{intra}} \): average learned interaction strength between nodes within the same cluster
    \item \textbf{Inter-community mean} \( \mu_{\text{inter}} \): average learned interaction strength between nodes across different clusters
\end{itemize}

\begin{figure*}[ht]
    \centering

    \begin{subfigure}[t]{0.48\textwidth}
        \includegraphics[width=\textwidth]{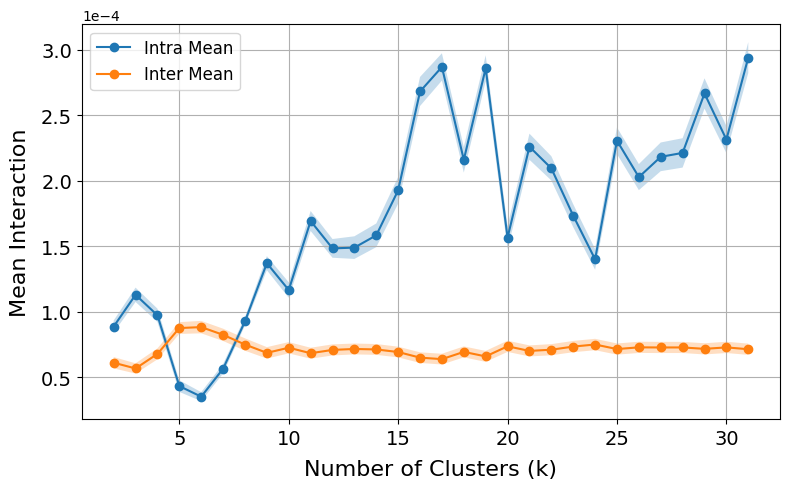}
        \caption{Interaction 1}
    \end{subfigure}
    \hfill
    \begin{subfigure}[t]{0.48\textwidth}
        \includegraphics[width=\textwidth]{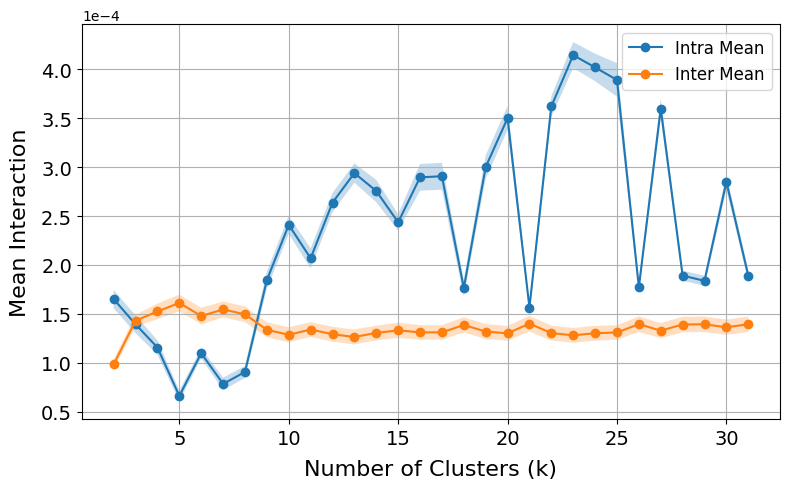}
        \caption{Interaction 2}
    \end{subfigure}
    \hfill
    \begin{subfigure}[t]{0.48\textwidth}
        \includegraphics[width=\textwidth]{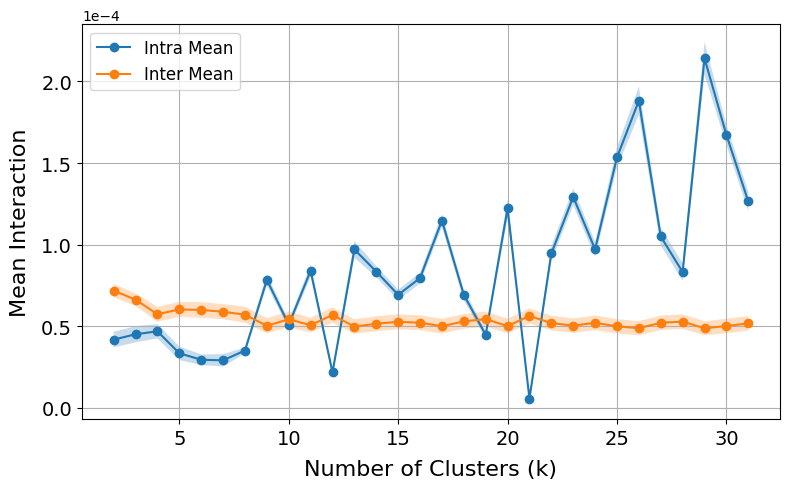}
        \caption{Interaction 3}
    \end{subfigure}


    \begin{subfigure}[t]{0.48\textwidth}
        \includegraphics[width=\textwidth]{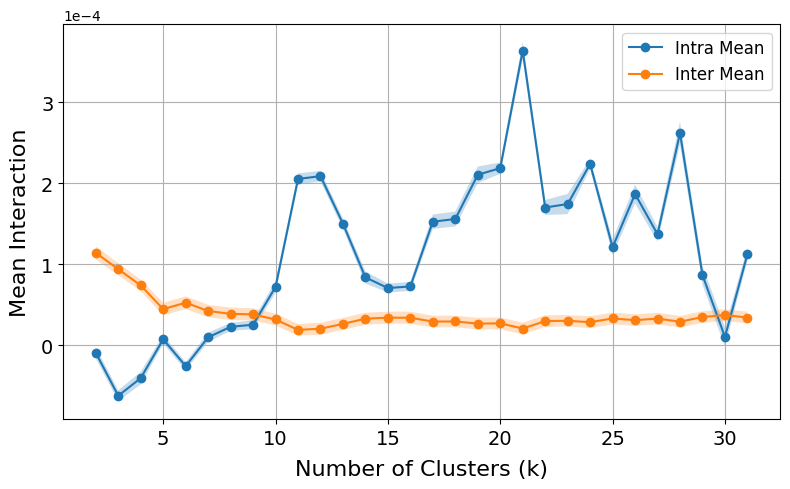}
        \caption{Interaction 4}
        \Description{Interaction 4}
    \end{subfigure}
    \hfill
    \begin{subfigure}[t]{0.48\textwidth}
        \includegraphics[width=\textwidth]{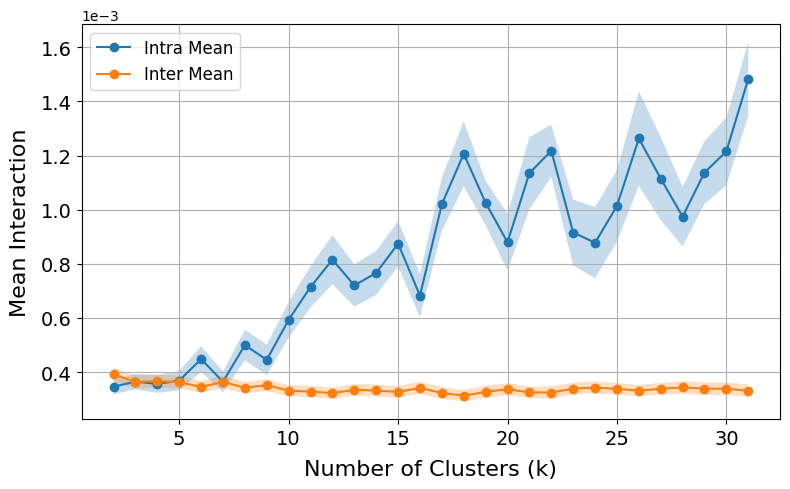}
        \caption{Interaction Aggregation}
        \Description{Interaction Aggregation}
    \end{subfigure}

    \caption{Intra- and inter-community statistics for each attention head computed using spectral clustering across varying numbers of clusters \( k \).}
    \Description{Intra- and inter-community statistics for each attention head computed using spectral clustering across varying numbers of clusters \( k \).}
    \label{fig:spectral_all_heads}
\end{figure*}

Figure \ref{fig:spectral_all_heads} shows intra and inter community statistics for each attention head computed using spectral clustering \cite{ng2002spectral}. The positive contrast ${\mu_{\text{intra},k} - \mu_{\text{inter},k}}$ shows that the interaction matrix is more aligned towards a community structure.

\subsection{Training Efficiency Analysis}
\label{appendix:train_time}

Table~\ref{tab:intergat_runtime} reports runtime breakdowns for InterGAT-GRU across two forecasting horizons (15-minute and 60-minute) on the SZ-Taxi dataset. The model exhibits consistent training efficiency, with average per-epoch durations remaining very similar across horizons.

\subsection{Training \& Validation Losses}
\label{appendix:training}

\begin{figure*}[ht]
\centering
\begin{subfigure}{0.6\textwidth}
    \includegraphics[width=\linewidth]{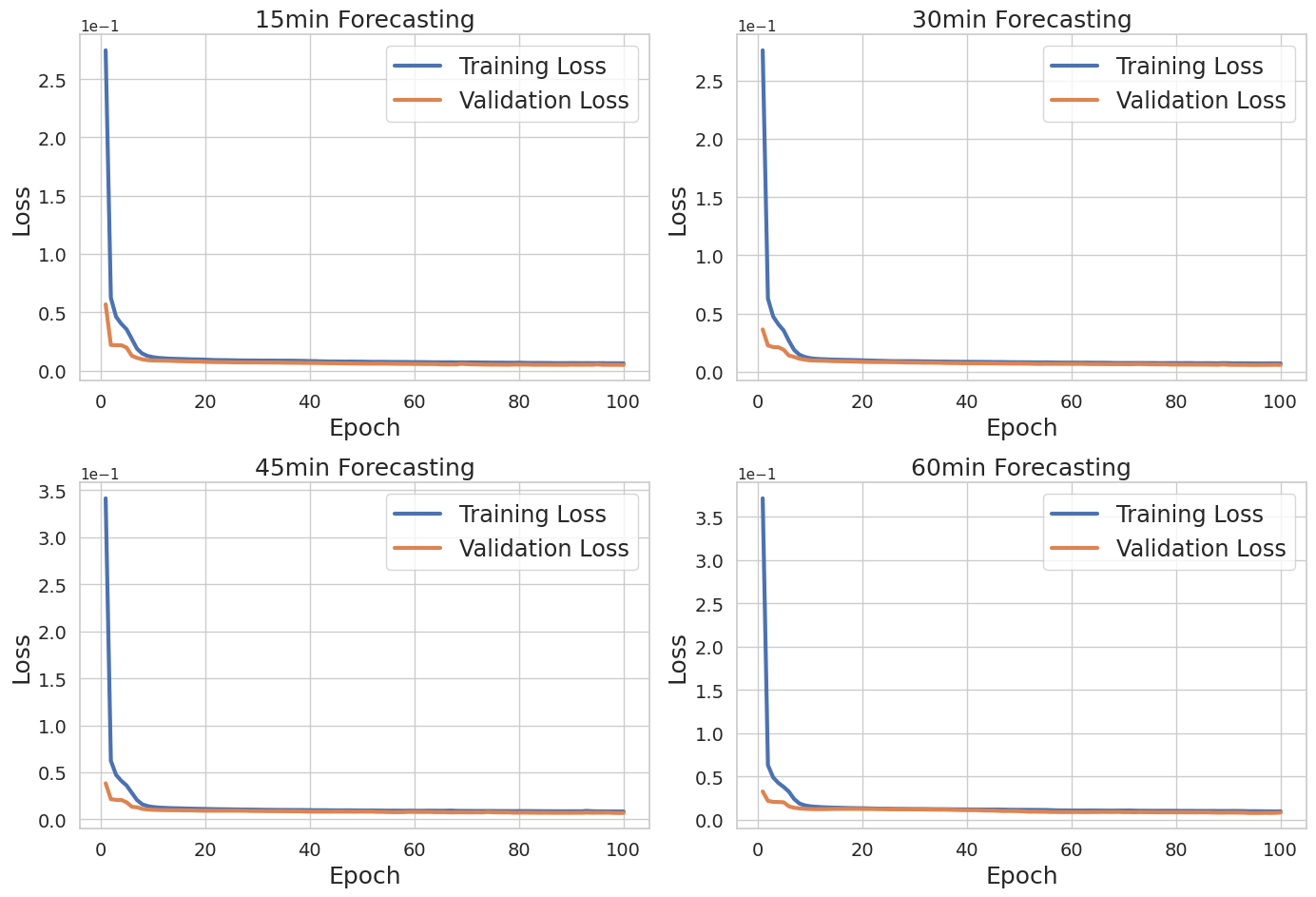}
    \caption{Los Loop: InterGAT-GRU training and validation loss across 15min, 30min, 45min, and 60min forecasting horizons.}
    \Description{Los Loop: InterGAT-GRU training and validation loss across 15min, 30min, 45min, and 60min forecasting horizons.}
    \label{fig:lossloop_forecasting}
\end{subfigure}
\hfill
\begin{subfigure}{0.6\textwidth}
    \includegraphics[width=\linewidth]{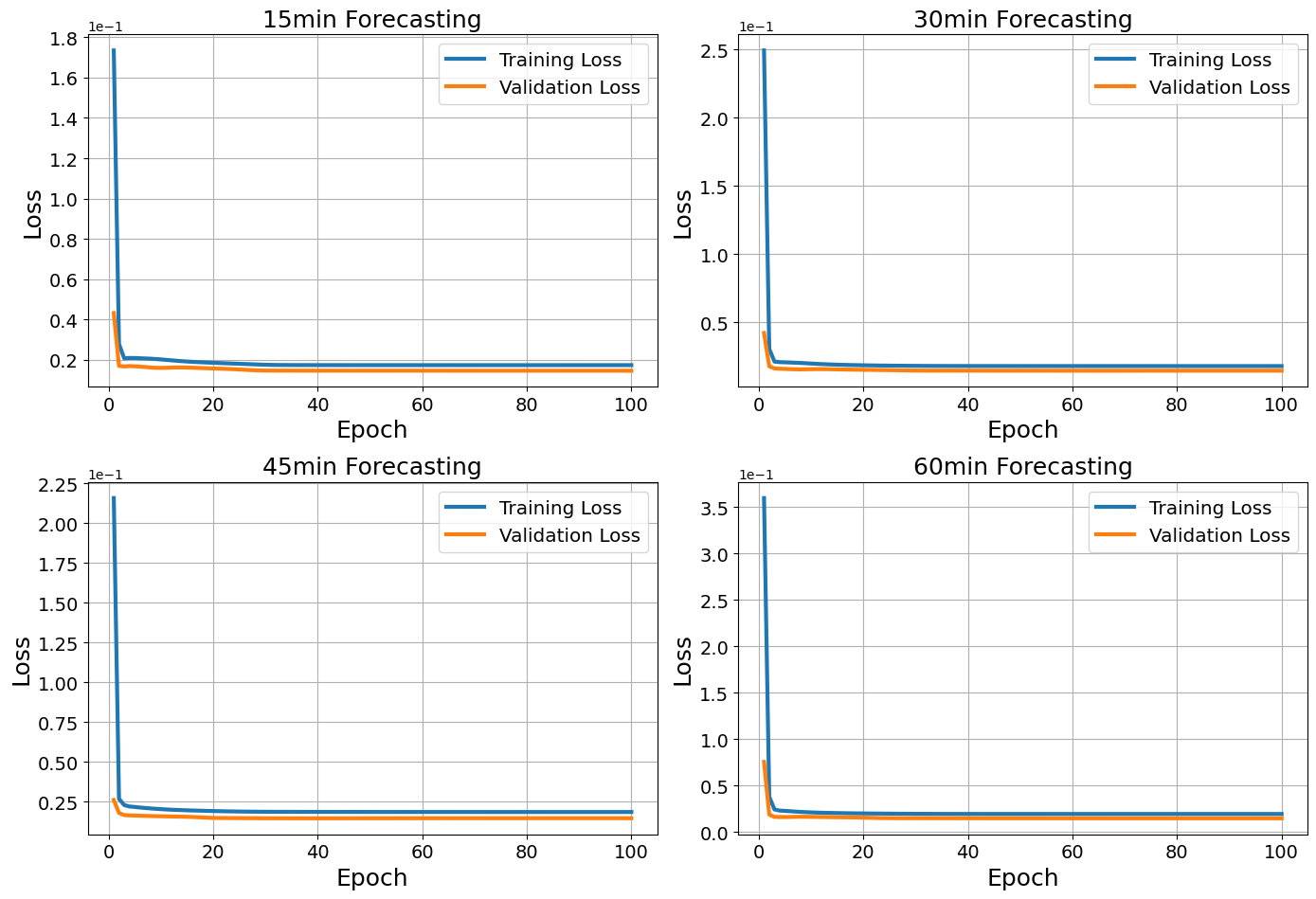}
    \caption{SZ-Taxi: InterGAT-GRU training and validation loss across 15min, 30min, 45min, and 60min forecasting horizons.}
    \Description{SZ-Taxi: InterGAT-GRU training and validation loss across 15min, 30min, 45min, and 60min forecasting horizons.}
    \label{fig:sztaxi_forecasting}
\end{subfigure}
\caption{Comparison of training and validation loss curves on two datasets using the InterGAT-GRU model.}
\Description{Comparison of training and validation loss curves on two datasets using the InterGAT-GRU model.}
\label{fig:forecast_loss_comparison}
\end{figure*}

We plot the training and validation loss curves of InterGAT-GRU across 100 epochs to evaluate its convergence behavior and generalization capacity. As shown in Figure~\ref{fig:forecast_loss_comparison}, the model exhibits stable and smooth convergence, with a rapid initial drop in both training and validation losses.

The validation loss consistently decreases and flattens early, indicating that the model is not overfitting and is able to generalize well to unseen data. The small gap between training and validation losses further suggests that the regularization mechanisms (e.g., L1 sparsity on the interaction matrix) are effective in promoting robust learning.

These trends confirm that InterGAT-GRU not only converges efficiently but also maintains strong generalization performance across epochs.

\begin{figure*}[!htbp]
  \centering
  \includegraphics[width=\textwidth]{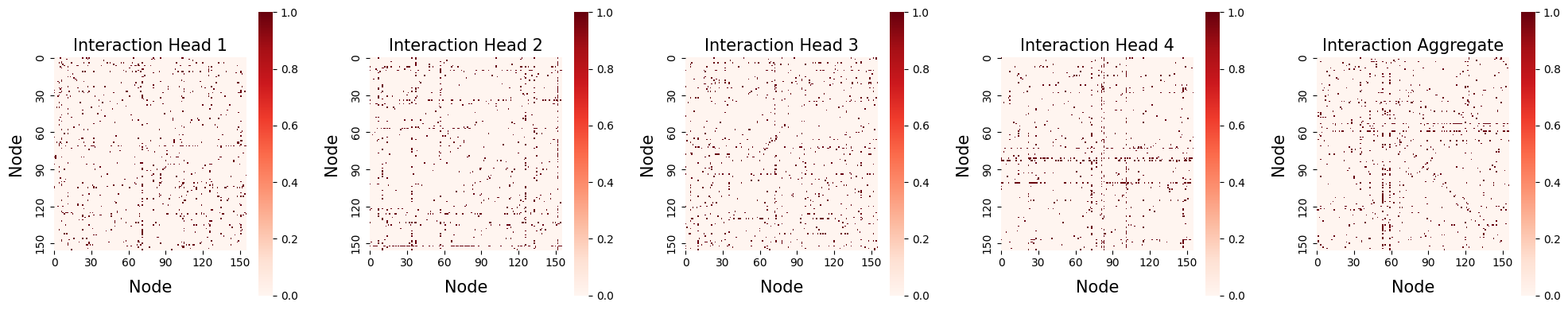}
  \caption{Interaction Matrix Heatmap - Binarized top 2\% of nodes}
  \Description{Interaction Matrix Heatmap - Binarized top 2\% of nodes.}
  \label{fig:interaction_heatmaps}
\end{figure*}

\subsection{Evaluation Metrics}
\label{appendix:evaluation}

To assess the prediction performance of our model, we use five standard evaluation metrics that compare the predicted traffic values $\hat{Y}_t$ with the ground truth $Y_t$ at time $t$:
\begin{enumerate}
    \item \textbf{Root Mean Squared Error (RMSE):}
    \[
    \text{RMSE} = \sqrt{ \frac{1}{n} \sum_{i=1}^{n} (Y_t^i - \hat{Y}_t^i)^2 }
    \]
    where $Y_t^i$ is the $i^{th}$ sample of $Y_t$, and $n$ is the total number of predictions (nodes $\times$ time steps $\times$ batches, if flattened). This metric penalizes large errors more heavily and is sensitive to outliers.

    \item \textbf{Mean Absolute Error (MAE):}
    \[
    \text{MAE} = \frac{1}{n} \sum_{i=1}^{n} \left| Y_t^i - \hat{Y}_t^i \right|
    \]
    MAE measures the average magnitude of errors, regardless of direction.

    \item \textbf{Accuracy:}
    \[
    \text{Accuracy} = 1 - \frac{ \| Y - \hat{Y} \|_F }{ \| Y \|_F }
    \]
    Accuracy measures how close the predicted matrix is to the ground truth in terms of Frobenius norm similarity.

    \item \textbf{Coefficient of Determination ($R^2$):}
    \[
    R^2 = 1 - \frac{ \sum_{i=1}^{n} (Y_t^i - \hat{Y}_t^i)^2 }{ \sum_{i=1}^{n} (Y_t^i - \bar{Y}_t)^2 }
    \]
    This metric evaluates how well the model explains the variance of the true data.

    \item \textbf{Explained Variance Score (Var):}
    \[
    \text{Var} = 1 - \frac{ \text{Var}(Y - \hat{Y}) }{ \text{Var}(Y) }
    \]
    It measures the proportion of variance in the target variable that is captured by the prediction.
\end{enumerate}

Lower values of RMSE and MAE indicate better predictive accuracy in terms of absolute error. Higher values of Accuracy, $R^2$, and Variance Score indicate that the model is capturing the underlying structure of the data more effectively.

\subsection{Interaction Matrix Visualization}
\label{appendix: int_matrix}

Figure~\ref{fig:interaction_heatmaps} shows the interaction matrices from each attention head and the aggregated output. Each matrix is min-max normalized to the \([0, 1]\) range, and only the top 2\% of values are retained to highlight the strongest learned interactions. This visualization reveals the most salient structural patterns captured by the model.

\end{document}